\documentclass{article}

\usepackage[nonatbib,final]{neurips_2022}

\usepackage[utf8]{inputenc} 
\usepackage[T1]{fontenc}    
\usepackage{hyperref}       
\usepackage{url}            
\usepackage{booktabs}       
\usepackage{amsfonts}       
\usepackage{nicefrac}       
\usepackage{microtype}      
\usepackage{xcolor}         
\usepackage{multirow}
\usepackage{enumitem} 
\usepackage{amsmath}
\usepackage{amssymb}
\usepackage{mathtools}
\usepackage{amsthm}
\usepackage{cite}
\usepackage{algorithm}
\usepackage{algorithmic}

\title{Data-Free Quantization with Accurate Activation Clipping and Adaptive Batch Normalization}

\author{%
  Yefei He, Luoming Zhang, Weijia Wu, Hong Zhou \\
  College of Biomedical Engineering \& Instrument Science\\
  Zhejiang University\\
  Hangzhou, China\\
  \texttt{\{billhe,zluoming,weijiawu\}@zju.edu.cn, zhouh@mail.bme.zju.edu.cn} \\
}

\begin{document}

\maketitle

\begin{abstract}
Data-free quantization compresses the neural network to low bit-width without access to original training data. Most existing data-free quantization methods cause severe performance degradation due to inaccurate activation clipping range and quantization error, especially for low bit-width. In this paper, we present a simple yet effective data-free quantization method with accurate activation clipping and adaptive batch normalization. Accurate activation clipping (AAC) improves the model accuracy by exploiting accurate activation information from the full-precision model. Adaptive batch normalization (ABN) firstly proposes to address the quantization error from distribution changes by updating the batch normalization layer adaptively. Extensive experiments demonstrate that the proposed data-free quantization method can yield surprisingly performance, achieving $\mathbf{64.33\%}$ top-1 accuracy of 4-bit ResNet18 on ImageNet dataset, with $\mathbf{3.7\%}$ absolute improvement outperforming the existing state-of-the-art methods.
\end{abstract}

\section{Introduction}
\label{introduction}
Deep learning has achieved breakthrough successes in many fields, such as computer vision\cite{yolov4, liu2022convnet} and natural language processing\cite{2018BERT}. Over-parameterization is an obvious feature of deep learning models compared to traditional methods. Since its birth, deep learning models have the disadvantages of massive parameters and high computational complexity. For example, ResNet18\cite{2016Deep} network has 11.7M parameters and 1.8GFLOPS calculations. These shortcomings limit the application of deep learning models in edge devices such as mobile phones. To solve this problem, model compression methods such as model quantization, distillation, and pruning have emerged in recent years. Among them, model quantization is one of the most commonly used compression methods. Quantization refers to mapping model parameters or activations from floating-point numbers to integers according to certain rules, thereby greatly reduce the size of the model and accelerate the inference process.

Quantizing a full-precision model to low bit-width directly often leads to a severe accuracy drop. To address this, two quantization methods, \textit{i.e.}, quantization-aware-training (QAT) and post-training quantization (PTQ) are proposed. The former aims to retrain the quantized model and consumes lots of computation resources. The latter directly quantize the pre-trained floating-point model and uses part of the training data to calibrate it. Both methods are data-driven that require real data in their quantization process. However, in many practical scenarios, the training dataset is not available for privacy policy or security issues.

Fortunately, data-free quantization can compress models without accessing any original training data. Many excellent prior studies\cite{2020ZeroQ, 2020Generative} try to solve the task by the generative methods with synthetic data. ZeroQ\cite{2020ZeroQ} and DSG\cite{DSG} proposed to generate fake data from the distribution~(\textit{i.e.}, mean and standard deviation) of BN layers. After using the fake data to update the range of quantized activations, the method achieved competitive result compared to the previous QAT methods at 8-bit, but the performance degrades significantly at low bit-width (especially 4-bit or lower). Although the synthetic data conforms to the distribution of BN layers, the activation range is actually determined by the peak value and there is no necessary connection between the extreme values of the data and its distribution. Therefore, we argut that the updated activation clipping range is inaccurate. 
Besides, the quantization process inevitably brings deviations to both weights and activations, thus disturbing the distribution of intermediate feature maps. The distribution mismatch between the feature maps and the fixed BN statistics will further aggravate the performance decrease. However, this problem is disregarded by previous data-free quantization methods and how to alleviate this mismatch remains an open question.
Furthermore, recent studies \cite{2020Generative, choi2021qimera, zhong2021intraq} proposed neural network-based data generators to fine-tune the quantized model. After fine-tuning, their work achieved better results than \cite{2020ZeroQ} when models were quantized to 4-bit. Nevertheless, training a generator requires a lot of time and computing resources, and fine-tuning can be less useful or even harmful for 8-bit quantization. How to 
quantize models generally and efficiently is still a problem.

To address the above issues, we present a simple yet effective data-free quantization method with two core components, \textit{i.e.}, accurate activation clipping and adaptive batch normalization. Accurate activation clipping uses the classification results of the teacher model and an absolute value loss function to optimize the synthetic data, the generated data contains accurate activation range information of the original dataset. Adaptive batch normalization takes the shift of quantized feature map distribution into consideration. After determining the activation value range, we propose to adaptively update the BN statistics of the quantized model, alleviating the distribution mismatch. Finally, we propose to perform fine-tuning only when necessary to further improve accuracy. Even with limited computing resources that can't perform fine-tuning, our method can achieve competitive performance compared to fine-tuning-based method.

We summarize our main contributions as follows:
\begin{itemize}[leftmargin=10pt]
    \item We rethink the determinants of activation range and propose a novel method to generate data specifically for accurate activation clipping. This provides a new approach for extracting dataset's activation information from full-precision models.
    \item An adaptive batch normalization is proposed to alleviate the disturbance from distribution mismatch of BN layers.
    \item Combining the above methods with optional fine-tuning, we get a brand new three-step data-free quantization pipeline. Extensive experiments on the large scale ImageNet dataset prove that quantization with our method can surpass the state-of-the-art methods with $\mathbf{3.7\%}$ absolute improvement on 4-bit ResNet18.
\end{itemize}

\begin{figure*}[t]
\vskip 0.2in
\begin{center}
\centerline{\includegraphics[width=\columnwidth]{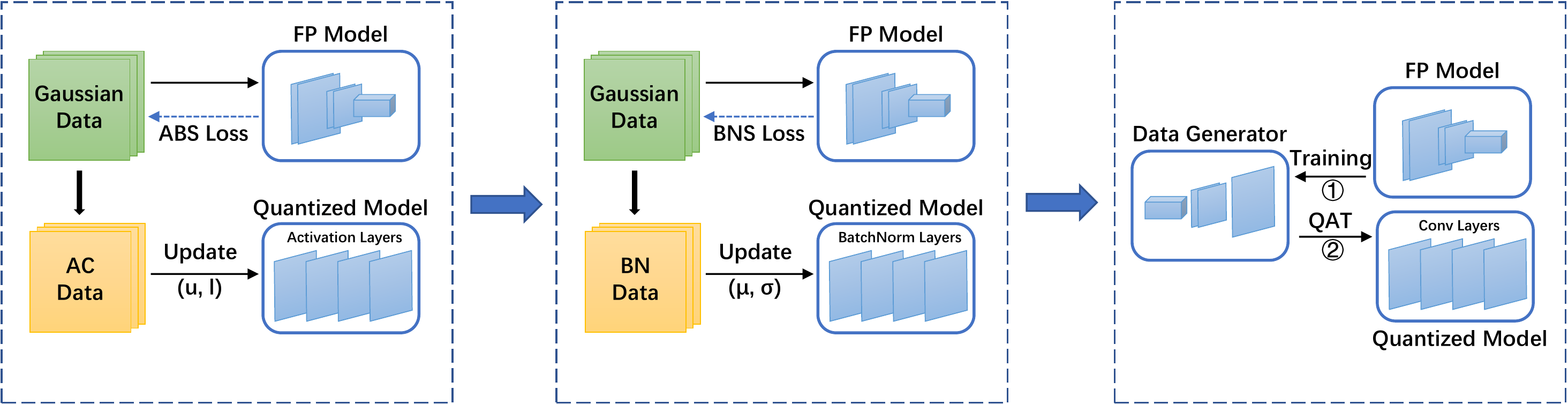}}
\caption{Proposed three-step pipeline for data-free quantization. Here, "AC" means activation clipping and "FP Model" denotes full-precision model.}
\label{pipeline}
\end{center}
\vskip -0.2in
\end{figure*}

\section{Related Work}
Data-free quantization was first proposed by DFQ\cite{2019DFQ} and quickly became the focus of research. There are two important questions in the field of data-free quantization: how to generate synthetic data and how to apply these data to improve the quantized model. Updating gaussian input using gradient backpropagation is a wildly adopted approach. Methods like ZeroQ\cite{2020ZeroQ} and DSG\cite{DSG} followed this scheme and extract information from BN layers to generate synthetic data for activation clipping. In this case, the key problem is how to recover the activation information of the training dataset from the full-precision model. Another common approach uses a network-based generator to synthesize data\cite{2020Generative, zhong2021intraq, choi2021qimera}. After generating data, they fine-tuned the quantized model to improve the accuracy. However, this can be much more time-consuming than the first approach.

To further improve the limit of data-free quantization, we argue that the method of generating data should be considered corporately with the method of using it. In other words, questions one and two should be considered together. For activation clipping, we analyzed the cause of the peak value of activations and proposed a new loss function $\mathcal{L}_\mathrm{ABS}$ that can generate data dedicated to activation clipping. Meanwhile, we propose a new method to apply the synthetic data, \textit{i.e.}, updating the BN layers’ statistics to alleviate the distribution mismatch. Both methods are fast and effective, and can be further combined with fine-tuning-based methods.

\section{Method}

\begin{figure}[ht]
\vskip 0.2in
\begin{center}
\centerline{\includegraphics[width=\columnwidth]{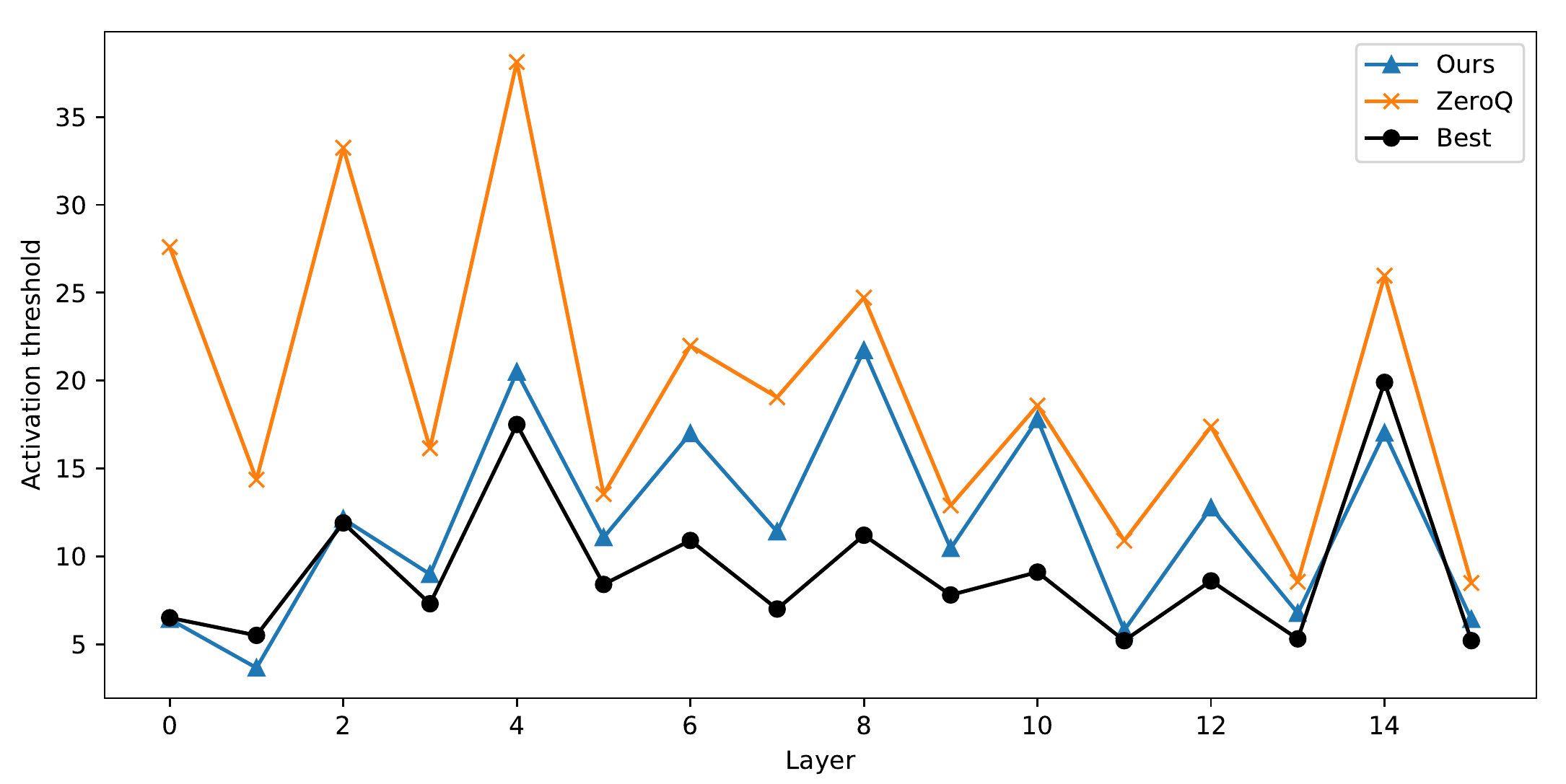}}
\caption{Activation range comparison for ResNet18 on ImageNet when quantized to 4-bit. Among them, the best clipping range is obtained by iterate through all values between $[0,50]$ (which basically covering all activation values) while inputting real dataset into the model.}
\label{activation_comparison}
\end{center}
\vskip -0.2in
\end{figure}

\subsection{Preliminaries} \label{preliminary}
We use uniform quantization in our study and experiments. For uniform quantization, given the bit-width $b$ and clip range $[l, u]$ for weight or activation, quantization-dequantization process as follows:

\begin{equation}
    \Delta = \frac{u-l}{2^b-1}  \label{delta}
\end{equation}
\begin{equation}
    Q(x) = \operatorname{round}(\frac{x-l}{\Delta})
\end{equation}
\begin{equation}
    D(x) = Q(x) * \Delta + l
\end{equation}

where $\Delta$ is the interval length,$Q(x)$ is the quantized representation of the data and $D(x)$ is the result of dequantization process of a value $Q(x)$.

There will be two main problems when it comes to data-free quantization. First, since the weight has been trained, the range of the weight is its minimum/maximum value. However, the clip range for activations of each layers depends on the specific input and is still unknown. Second, the statistics ($\mu$ and $\sigma$) of BN layers depend on the input and feature maps of the network, and have been fixed in the model. However, quantizing the model may shift the distribution of the intermediate feature maps, which consequently become inconsistent with the BN statistics. 

In this section, we propose two approaches, \textit{i.e.}, accurate activation clipping and adaptive batch normalization to address the corresponding two problems, which can achieve remarkable results in a short time. In low-bit quantization, we further use fine-tuning to improve the accuracy, thus having a three-step pipeline for data-free quantization, as shown in Figure~\ref{pipeline}.

\subsection{Accurate Activation Clipping} \label{accurate data}
To determine the clipping range of activations (\textit{i.e.} $u$ and $l$ in Equation~\eqref{delta}), one common way is to generate synthetic data and conduct a forward propagation with it. The peak value of activations is stored as the clipping range parameter. While previous studies\cite{2020ZeroQ, 2020Generative} generate fake data with the distribution of BN layers, it only provides a coarse prediction for the activation range. We argue that it is not the optimal choice for updating the clipping range because the peaks are not directly related to the distribution of the data. This is illustrated in Figure~\ref{activation_comparison}, where we plot the best activation clipping range and the data-determined range for every layer in the ResNet18 model. With distribution-consistent data from BN layers, ZeroQ\cite{2020ZeroQ} presents a dissatisfied performance compared to the optimal activation value. So what exactly determines the activation clipping range?

\begin{figure}[ht]
\vskip 0.2in
\begin{center}
\centerline{\includegraphics[width=\columnwidth]{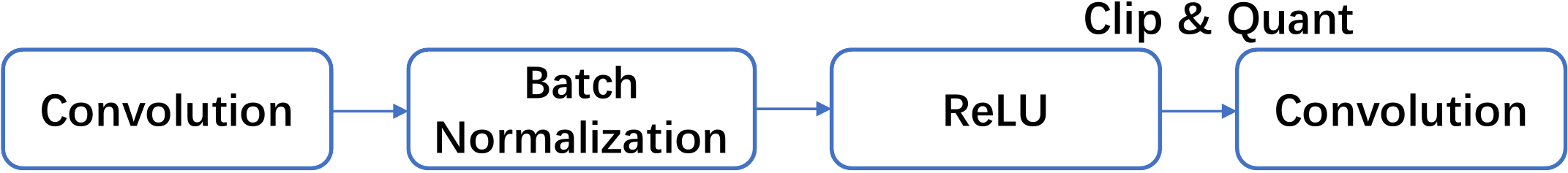}}
\caption{A common block structure of quantized CNNs}
\label{common_block}
\end{center}
\vskip -0.2in
\end{figure}

Figure~\ref{common_block} presents a common block structure of quantized CNNs. Before clipping, ReLU takes feature map of convolution layers as input. Thus, the lower bound $l$ is zero without doubt, while the upper bound $u$ depends on the maximum value of feature map. The recent study~\cite{2016CAM} revealed that feature map is the response of the network to category features where high response is related to significant features and low response is related to irrelevant features such as background. Therefore, to emulate the response of the real dataset, synthetic data should make the network to be highly responsive.

To this end, we propose a novel method to generate accurate activation clipping data. We notice that deep learning models perform image classification tasks on large-scale datasets well with rich category-related information. Thus we try to let the model to learn the most responsive data by itself. For example, given a target label "flowers" and a gaussian-random image, if we input the gaussian image to the network directly, the result is unlikely to classify it as a flower. However, if we have an appropriate loss function, calculate the loss according to the target label and backpropagate to the image, the confidence of classifying as "flowers" will increase as we iterate. Finally, the network generates the "most flower-liked" image adaptively. With the high-quality generated image used as input, the model will emerge higher response and emulate the real dataset's activation peaks. In other words, we have to solve the following optimization problems:
\begin{equation}
    \min_{x} \mathbb{E}_{x, y}
    \left[\ell({M}(x), y)\right],
\end{equation}

where $x$ is the synthetic input data, $y$ is the target label, $\ell$ is the loss function and $M$ is the full-precision model.

\begin{figure}[ht]
\vskip 0.2in
\begin{center}
\centerline{\includegraphics[width=\columnwidth]{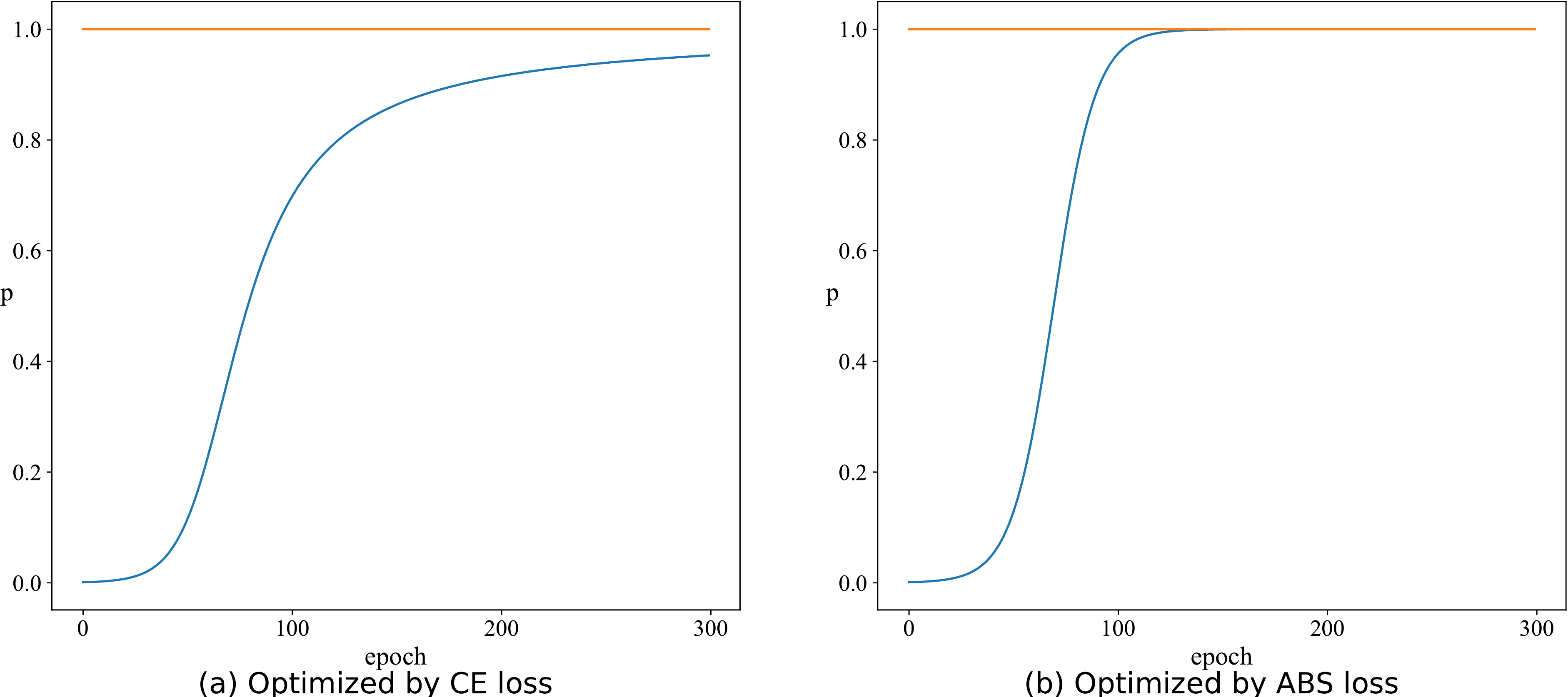}}
\caption{Toy experiment of optimization using different loss functions. We take M(x)=x as the model and optimize the input. $p$ represents prediction probability of target class.}
\label{effectofabs}
\end{center}
\vskip -0.2in
\end{figure}

To select an appropriate loss function, we take a deeper look at this problem. The Cross-Entropy (CE) loss is the most commonly used loss function in image classification tasks and it’s a good indicator of how well the model classify input images. Therefore, we use the CE loss to judge the quality of the generated data, where lower loss means better generated data. However, when it comes to backpropagation, CE loss is not the best choice. A toy example is used to illustrate the problem, as shown in Figure~\ref{effectofabs}(a). We can see that using CE loss for backpropagation cannot optimize the loss function to the optimum. Details of the toy experiment can be found in Appendix~\ref{sec:toy}.

The derivative computed by CE loss can be formulated as:
\begin{gather}
    \frac{\partial \mathcal{L}_\mathrm{CE}}{\partial M_y(x)}=\frac{\partial \mathcal{L}_\mathrm{CE}}{\partial p_y}\cdot\frac{\partial p_y}{\partial M_y(x)}\\ 
    =-\frac{1}{p_y}\cdot p_y \cdot (1-p_y)\\ \label{dsoftmax}
    =p_y - 1
\end{gather}
where $p_y$ is the probability of classify to label y (\textit{i.e.}, the result of $\operatorname{softmax}$ operation). With the increase of $p_y$, the derivative used for backpropagation will be smaller, which makes the optimization process saturated before it reaches the optimal state. 

It can be seen from Equation~\eqref{dsoftmax} that the problem is mainly caused by $softmax$ operation, where all the scores are brought into calculation. Actually, in this peak value problem, we only care about the score corresponding to the target label and hope to reflect it on the input image. Scores of other labels or their appearance on the input image are irrelevant to the model’s response. Therefore, we propose a hard-label absolute value loss (ABS Loss):
\begin{equation}
    \mathcal{L}_\mathrm{ABS} = - {M}_y(x)
    \label{absloss} 
\end{equation}

Thus, only the regions that the model considers to have an impact on the "flower" features will be enhanced, and the intermediate feature map will also produce a higher response. Using Equation~\eqref{absloss} as loss functions in the algorithm also make the Cross-Entropy loss drop to 0 quickly, making the data generation process even faster (see Figure~\ref{effectofabs}(b)). It can also be seen from the Figure~\ref{activation_comparison} that the data we generate is very close to the optimal value of the activation clipping range.



\subsection{Adaptive Batch Normalization}

\begin{figure}[ht]
\vskip 0.2in
\begin{center}
\centerline{\includegraphics[width=\columnwidth]{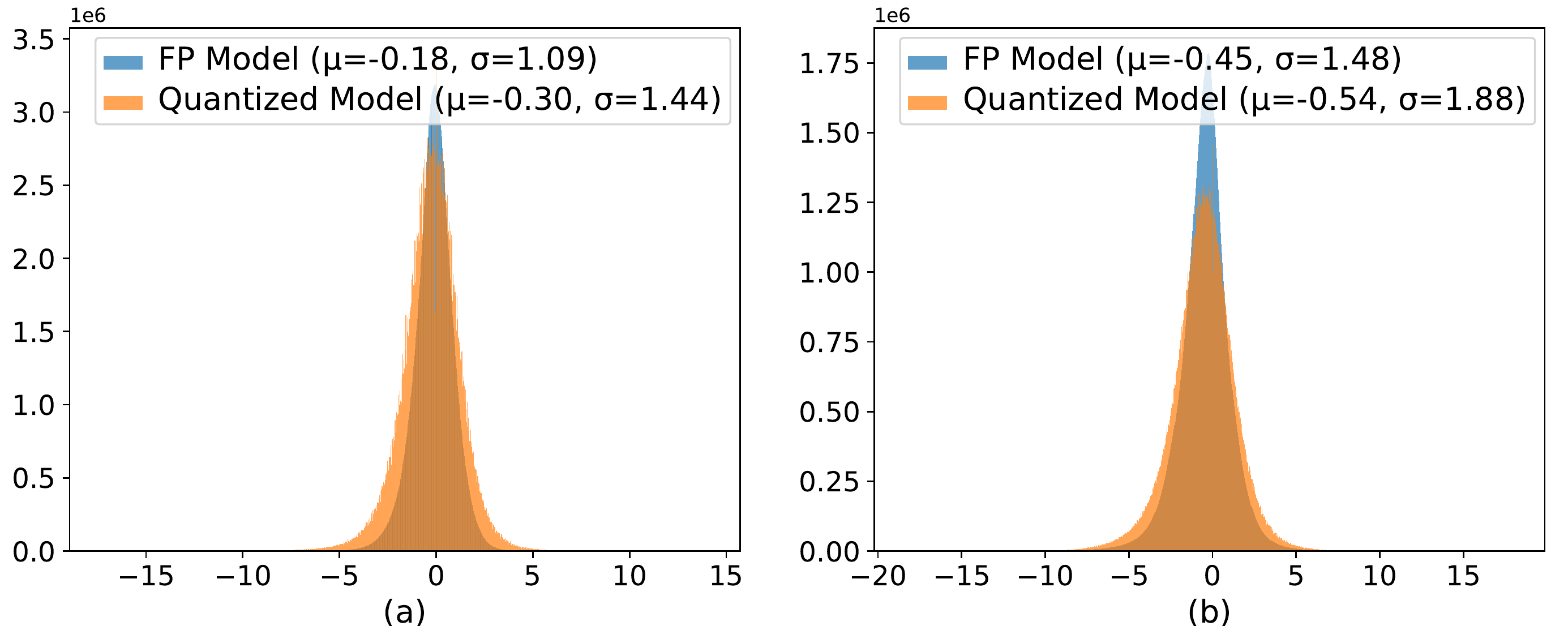}}
\caption{Comparison between featuremaps of original model and quantized model. (a): layer1.1.conv2, (b): layer2.0.conv2.}
\label{quantized_distribution}
\end{center}
\vskip -0.2in
\end{figure}

Previous research\cite{2019DFQ} has found that when the convolutional layer is quantized, the output featuremap distribution will have a certain offset compared to the original model. When we quantize models to low bit-width, this phenomenon will be more obvious, as shown in Figure~\ref{quantized_distribution}. In this case, since the distribution of the input data does not match the statistic stored in the BN layer, the BN layer not only fails to normalize the data, but may also have a side effect that harms the model accuracy. This statistic inconsistency is caused by quantization errors. When we have the training dataset, we can use QAT to retrain the weights and BN layers of the network. However, in the case of data-free quantization, we do not have training data to optimize the quantization error. Hence, in this study we look at this problem from another perspective. Since the model has a large number of weight parameters, it's difficult to update without training data. But the BN layers contains few parameters and only depends on the distribution of the input data. In order to alleviate the distribution mismatch and improve the accuracy of the quantized model, we choose to update the statistics of BN layers and let BN layers adapt to the errors caused by quantization.

Although the data generated in section \ref{accurate data} can restore the activation range very well, it is not suitable for updating the statistics of the BN layer. As analyzed above, updating the activation range is an extremum problem while updating the BN statistics is a data distribution problem. Therefore, we use the distribution-consistent data proposed in \cite{2020ZeroQ} to update the BN statistics.

The BN statistics (BNS) loss function is defined as follows:
\begin{gather}
    \mathcal{L}_\mathrm{BNS} = \sum_{i=0}^L{\Vert \widehat{\mu_i}- \mu_i \Vert_2}^{2} + {\Vert \widehat{\sigma_i}- \sigma_i \Vert_2}^{2}
    \label{BNSloss}
\end{gather}
Among them, $\mu_i$, $\sigma_i$ are the mean and standard deviation information stored in the i-th BN layer in the full-precision model. $\widehat{\mu_i}$ and $\widehat{\sigma_i}$ are the mean and standard deviation of input data at layer $i$. Similar to AAC, the random input image is backpropagated through the loss function \eqref{BNSloss}, and the generated data obtained contains the distribution information of the training dataset. By feeding this data into the network, we can measure the distribution disturbance and update the BN statistics well.

\subsection{Three-Step Pipeline}
In the case of low bit-width (especially 4-bit or lower) quantization, it is difficult to achieve ideal results without fine-tuning the weights. Methods like GDFQ\cite{2020Generative} propose to use a network-based generator to generate fake data and perform retraining on the quantized model. 

However, as analyzed in section\ref{preliminary}, determining activation range and BN statistics are key problems in data-free quantization. While retraining the weights improve the accuracy, these two problems are neglected by previous fine-tuning methods. Meanwhile, our corresponding methods proposed above are effective and fast that can be completed within one minute. Therefore, we use our method as the basis for fine-tuning, which makes up for their shortcomings and greatly improves the accuracy. At this point, we get a three-step data-free quantization pipeline, as shown in Figure~\ref{pipeline}.

Fine-tuning can consume lots of time and computing resources while our method are efficient. Therefore, we try to seek a balance between time and effect. In the case of 8-bit, our two-step method (without fine-tuning) has been able to achieve ideal results, and the improvement brought by fine-tuning is subtle. In the case of low bit-width, although the improvement brought by fine-tuning is still considerable and a complete three-step method is worth performing, the two-step method achieves a huge improvement over recent study DSG\cite{DSG} and is comparable with fine-tuning based method, which will be further analyzed in section~\ref{sotasection}.

\section{Experiments} \label{experiment}
\subsection{Implementation Details} \label{implementation details}
This research is based on the Pytorch framework and all experiments are conducted using RTX3090 GPUs. For the generation of accurate activation clipping data, we used a learning rate of 0.2 to optimize 200 iterations; for the generation of distribution-consistent data, we used a learning rate of 0.5 to optimize 500 iterations. For data generator, we follow the settings introduced by \cite{2020Generative}. To make a fair comparison, the method of quantizing models is the same as \cite{2020ZeroQ} and \cite{2020Generative}, where all the convolution layers and linear layers are quantized to the same bit-width.

We conduct all experiments on the large-scale benchmark dataset ImageNet (ILSVRC12)\cite{2009ImageNet}. In order to verify the versatility of our method, we tested several commonly used network structures, including ResNet\cite{2016Deep}, VGG\cite{2014VGG}, Inception v3\cite{2016Inception} and MobileNetV2\cite{2018Mobilenetv2}. All networks use the pre-trained model provided by pytorchcv\cite{you2019torchcv}.

\subsection{Comparison with State-of-the-arts} \label{sotasection}

\begin{table*}[t] 
    \caption{Comparisons of different network architecture on ImageNet. Here, "No Data" means no training data is used, "No FT" stands for no retraining the weight, and "FP" means full-precision pretrained model.}
    \label{mainresult}
\centering
\begin{sc}
\begin{tabular}{cccccc}
\hline
Model & No Data & No FT & Method & Bit-width(W/A) & Acc.(\%)\\
\hline
\multirow{12}{*}{ResNet18}
& $\times$ & $\times$ & FP & 32/32 & 71.47 \\
& $\surd$ & $\surd$ & ZeroQ\cite{2020ZeroQ} & 4/4 & 26.04 \\
& $\surd$ & $\surd$ & DSG\cite{DSG} & 4/4 & 34.53 \\
& $\surd$ & $\surd$ & Ours & 4/4 & 55.06 \\
& $\surd$ & $\times$ & GDFQ\cite{2020Generative} & 4/4 & 60.60 \\
& $\surd$ & $\times$ & Ours & 4/4 & \textbf{64.33} \\
\cline{2-6}
& $\times$ & $\times$ & FP & 32/32 & 71.47 \\
& $\surd$ & $\surd$ & DFQ\cite{2019DFQ} & 8/8 & 69.70 \\
& $\surd$ & $\times$ & GDFQ\cite{2020Generative} & 8/8 & 70.80 \\
& $\surd$ & $\surd$ & ZeroQ\cite{2020ZeroQ} & 8/8 & 71.43 \\
& $\surd$ & $\surd$ & Ours & 8/8 & 71.43 \\
& $\surd$ & $\surd$ & DSG\cite{DSG} & 8/8 & \textbf{71.49} \\
\hline
\multirow{5}{*}{BN-VGG16}
& $\times$ & $\times$ & FP & 32/32 & 74.28 \\
& $\surd$ & $\surd$ & ZeroQ\cite{2020ZeroQ} & 4/4 & 1.15 \\
& $\surd$ & $\times$ & GDFQ\cite{2020Generative} & 4/4 & 67.10 \\
& $\surd$ & $\surd$ & Ours & 4/4 & 68.59 \\
& $\surd$ & $\times$ & Ours & 4/4 & \textbf{70.10} \\
\hline
\multirow{6}{*}{Inception V3}
& $\times$ & $\times$ & FP & 32/32 & 78.80 \\
& $\surd$ & $\surd$ & ZeroQ\cite{2020ZeroQ} & 4/4 & 26.84 \\
& $\surd$ & $\surd$ & DSG\cite{DSG} & 4/4 & 34.89 \\
& $\surd$ & $\surd$ & Ours & 4/4 & 69.20 \\
& $\surd$ & $\times$ & GDFQ\cite{2020Generative} & 4/4 & 70.39 \\
& $\surd$ & $\times$ & Ours & 4/4 & \textbf{72.74} \\
\hline
\multirow{7}{*}{MobileNet V2}
& $\times$ & $\times$ & FP & 32/32 & 73.03 \\
& $\surd$ & $\times$ & GDFQ\cite{2020Generative} & 4/4 & 59.24 \\
& $\surd$ & $\times$ & Ours & 4/4 & \textbf{59.50} \\
\cline{2-6}
& $\surd$ & $\surd$ & DFQ\cite{2019DFQ} & 8/8 & 72.33 \\
& $\surd$ & $\times$ & GDFQ\cite{2020Generative} & 8/8 & 72.88 \\
& $\surd$ & $\surd$ & ZeroQ\cite{2020ZeroQ} & 8/8 & 72.91 \\
& $\surd$ & $\surd$ & Ours & 8/8 & \textbf{72.93} \\
\hline
\end{tabular}
\end{sc}
\end{table*}

In this subsection, we evaluate our method by comparing it with existing state-of-the-arts data-free quantization methods over various network architecture. Since the accuracy in 8-bit quantization is very close to the full-precision model, we mainly report the results at 4-bit, as shown in Table\ref{mainresult}.

We can observe that for W4A4 (\textit{i.e.}, quantize both weights and activations to 4-bit), our method outperforms all other methods by large margins on different network architectures. This illustrates the effectiveness of our accurate activation clipping and BN statistics update. Even if our method does not use fine-tuning on many models, the effect is very close to or exceeds that of GDFQ, which requires fine-tuning. As we all know, fine-tuning consumes a lot of GPU time. Our approach (without FT, two-step) only takes about one minute to complete, strikes a better balance between time and results.

For W8A8, the effects of various methods are very close to the full-precision model. In this case, we find that if the generated data is used for fine-tuning, the effect is subtle or may even be worse than the result of not using fine-tuning. We recommend not using fine-tuning for 8-bit quantization.

\subsection{Ablation Study}
\subsubsection{Three-step pipeline}
To further verify the effects of each part of three-step pipeline, we used the ResNet18 model to perform ablation experiments on ImageNet dataset. Table~\ref{ablation} shows the performance under different settings. We can see that all three steps have played an important role in improving accuracy. When we combine them, the proposed three-steps pipeline obtained achieves the effect of SOTA, indicating that the improvements brought by the three steps can be superimposed. The three steps consider a better activation clipping range, updating BN statistics to adapt to the quantization error, and fine-tuning the weight separately, that's why our method can obtain better results. It is worth mentioning that when only the activation clipping range is updated, our performance greatly surpasses ZeroQ\cite{2020ZeroQ}, indicating that our method can better restore the activation value range of the real dataset.

\begin{table}[h]
\caption{Ablation study of three-step pipeline on 4-bit quantization. Here, "AC" donotes activation clipping, "BN Update" means updating the statistics in BN layers of quantized model and "FT" means using synthetic data to fine-tune the quantized model.}
\label{ablation}
\vskip 0.15in
\begin{center}
\begin{small}
\begin{sc}
\begin{tabular}{ccccr}
\toprule
With AC & With BN update & With FT & Acc.(\%) \\
\midrule
$\surd$ (ZeroQ)  & $\times$ & $\times$ & 26.04 \\
\midrule
$\times$  & $\times$ & $\times$ & 10.22 \\
$\surd$  & $\times$ & $\times$ & 43.03 \\
$\surd$  & $\surd$ & $\times$ & 55.06 \\
$\surd$  & $\surd$ & $\surd$ & \textbf{64.33} \\
\bottomrule
\end{tabular}
\end{sc}
\end{small}
\end{center}
\vskip -0.1in
\end{table}

What's more, we valid the effects of proposed method over several state-of-the-art fine-tuning-based methods. As shown in Table~\ref{tab:otherft}. In all cases, our method can further improve the accuracy, proving the robustness of our method and that an accurate activation range is an important basis for the fine-tuning-based method.

\begin{table}[h]
\caption{Results of 4-bit ResNet18 on ImageNet. The results are obtained with the official code from authors}
\label{tab:otherft}
\begin{center}
\begin{sc}
\begin{tabular}{cc}
\hline
Method               & Top-1 Acc.(\%)         \\ \hline
real data            & 67.90                  \\ \hline
GDFQ\cite{2020Generative}                 & 60.60                  \\
\textbf{GDFQ+Ours}   & \textbf{64.33 (+3.73)} \\ \hline
Qimera\cite{choi2021qimera}               & 62.98             \\
\textbf{Qimera+Ours} & \textbf{63.72 (+0.74)} \\ \hline
IntraQ\cite{zhong2021intraq}               & 66.47                  \\
\textbf{IntraQ+ours} & \textbf{67.06 (+0.59)} \\ \hline
\end{tabular}
\end{sc}
\end{center}
\end{table}

\subsubsection{Selection of Loss Function} \label{lossanalyzesection}
We further investigate the performance of generating accurate activation data using different loss functions. The ResNet18 model is quantized to 4-bit. We iterate on the gaussian image for 600 epochs to generate synthetic data, and use this data to update the activation clipping range. Table \ref{lossselect} shows the performance under different settings, where “MAE” denotes mean average error and "MSE" stands for mean square error. We feed the generated data into the model and calculate the Cross-Entropy loss to measure how well the model responds to the data.

\begin{table}[h]
\caption{Selection of different loss functions}
\label{lossselect}
\vskip 0.15in
\begin{center}
\begin{small}
\begin{sc}
\begin{tabular}{clll}
\hline
\multicolumn{1}{l}{Model} & Method        & Loss       & Acc.(\%)       \\ \hline
\multirow{5}{*}{ResNet18} & ABS+BNS       & 2.86       & 28.46          \\
                          & Cross-entropy & 1.77       & 33.68          \\
                          & MAE           & 1.10       & 41.13          \\
                          & MSE           & 0.25       & 41.39          \\
                          & ABS           & \textbf{0} & \textbf{43.03} \\ \hline
\end{tabular}
\end{sc}
\end{small}
\end{center}
\vskip -0.1in
\end{table}

It can be seen that the data generated by the proposed ABS loss can make the model predict labels exactly as we expected. This also achieved the highest classification accuracy after updating the quantized model.

An intuitive idea is to allow the synthetic data to make the model highly responsive while conforming to the distribution determined by the BN statistics. However, in experiments we found that if we add the two loss functions together, the optimization process becomes unstable and is unable to converge.

\subsection{Quantization Efficiency}
In this subsection, we compare our two-step and three-step pipeline with the existing data-free quantization methods, as shwon in Table~\ref{speed}. Methods that update gaussian input to determine the activation range is way faster than methods with fine-tuning. Among them, our method achieves the best accuracy and brings 21\% absolute improvement on 4-bit ResNet18. This is very useful on devices that have limited computing power and cannot perform fine-tuning. When fine-tune is required, our method can greatly improve the accuracy while barely increasing the time consumption.

\begin{table}[h]
\caption{Comparison of time cost of 4-bit ResNet-18 with different data-free quantization methods.}
\label{speed}
\begin{center}
\begin{sc}
\begin{tabular}{llll}
\hline
Method         & With FT  & Acc.(\%) & Time (min) \\ \hline
ZeroQ          &$\times$  & 26.08    & 0.83        \\
DSG             &$\times$ & 34.53    & -        \\
Ours (two-step)  &$\times$ & 55.06    & 1.38         \\
GDFQ            &$\surd$ & 60.60    & 139.6       \\
Ours (three-step) &$\surd$ & 64.33    & 141.2       \\ \hline
\end{tabular}
\end{sc}
\end{center}
\end{table}

\section{Conclusion and Discussion}
In this paper, we propose two novel techniques: accurate activation clipping and adaptive batch normalization to improve the accuracy of data-free quantization. First, we analyze the origin of activation peak values and construct a new scheme to generate synthetic data that can restore the activation range of the original dataset, thus helping better quantizing the activations. Next, we analyze the mismatch between quantized featuremap distribution and statistics stored in BN layers. We propose to update the BN statistics adaptively and let the BN layers adapt to the errors caused by quantization. Finally, we combine the above two methods with an optional fine-tune module and get a three-step quantization pipeline, which can further improve the accuracy, allowing users to strike a balance between time consumption and results. Extensive experiments prove that our methods outperforms the existed data-free quantization methods by a large margin.

We also find that, since the existing methods mainly focus on updating the clipping range of activation values, they do not work well with models like MobileNet that do not use traditional $\operatorname{ReLU}$ as the activation function. How to  quantize networks like MobileNet effectively is a question worth studying in the future.

\bibliography{paper}
\bibliographystyle{abbrv}

\section{Appendix}
\subsection{Implementation details of toy experiment} \label{sec:toy}

In the toy experiment shown in Figure~\ref{effectofabs}, we let the model to be an identity transform. Given a target label, we calculate different loss functions and backpropagate to the input, which will make the value of target class higher during iteration. We run both experiments for 300 iterations. The algorithm is summarized below.

\begin{algorithm}[h]
  \caption{Toy experiment of optimizing by different loss functions}
  \label{clipdatagenerate}
\begin{algorithmic}
  \STATE {\bfseries Require:} Identity transform model $M$, a loss function $\mathcal{L}$, random data $\mathbf x \in\mathbb R^{n}$ and a target label $t \in [0, n-1]$.
  \FOR{$i=1$ {\bfseries to} $num\_iterations$}
  \STATE Forward propagate $M(\mathbf{x})=\mathbf{x}$
  \STATE Calculate loss value according to the choice of loss function $\mathcal{L}$
  \STATE Backward propagate and update $\mathbf{x}$
  \ENDFOR
  \STATE {\bfseries Return:} Target value $\mathbf{x}_t$
\end{algorithmic}
\end{algorithm}



\end{document}